\documentclass[letterpaper]{article} 
\usepackage{aaai2026} 
\usepackage{times} 
\usepackage{helvet} 
\usepackage{courier} 
\usepackage[hyphens]{url} 
\usepackage{graphicx} 
\usepackage{natbib} 
\usepackage{caption} 
\usepackage{algorithm}
\usepackage{algorithmicx}
\usepackage{algpseudocode}

\usepackage{times}
\usepackage{helvet}
\usepackage{courier}
\usepackage{xcolor}
\usepackage{booktabs}
\usepackage{colortbl}
\usepackage{multirow}
\usepackage{multicol}
\usepackage{listings}
\usepackage[table]{xcolor}
\usepackage{makecell}

\usepackage{fvextra}

\fvset{breaklines=true} 
\usepackage[most]{tcolorbox}
\tcbset{
  myboxstyle/.style={
    colback=gray!20,
    colframe=gray!140,
    sharp corners,
    boxrule=0.8pt,
    fonttitle=\bfseries,
    enhanced,
    beforeafter skip=0pt,  
    left=5pt, right=5pt, top=5pt, bottom=5pt,
  }
}

\usepackage{newfloat}
\usepackage{listings}
\usepackage{marvosym}
\DeclareCaptionStyle{ruled}{labelfont=normalfont,labelsep=colon,strut=off} 
\floatstyle{ruled}
\newfloat{listing}{tb}{lst}{}
\floatname{listing}{Listing}
\title{\methodname: A VLM-in-the-Loop Adversary for Enhancing Driving Policy Robustness}
\author{
  Qimao Chen\textsuperscript{\rm 1,\rm 3\equalcontrib}, Fang Li\textsuperscript{\rm 3\equalcontrib}, Shaoqing Xu\textsuperscript{\rm 2,\rm 3\equalcontrib\thanks{Project Leader.}}, Zhiyi Lai\textsuperscript{\rm 3}, Zixun Xie\textsuperscript{\rm 3,\rm 4}, Yuechen Luo\textsuperscript{\rm 1,\rm 3}, \\Shengyin Jiang\textsuperscript{\rm 3}, Hanbing Li\textsuperscript{\rm 3}, Long Chen\textsuperscript{\rm 3}, Bing Wang\textsuperscript{\rm 3}, Yi Zhang\textsuperscript{\rm 1\textsuperscript{\Letter}}, Zhi-Xin Yang\textsuperscript{\rm 2\textsuperscript{\Letter}}
}
\affiliations{
  \textsuperscript{\rm 1}Tsinghua University,
  \textsuperscript{\rm 2}University of Macau,
  \textsuperscript{\rm 3}Xiaomi EV,
  \textsuperscript{\rm 4}Peking University
  \\
  cqm24@mails.tsinghua.edu.cn
  
}

\newcommand{\methodname}{VILTA}

\begin{document}

\maketitle

\begin{abstract}
The safe deployment of autonomous driving (AD) systems is fundamentally hindered by the long-tail problem, where rare yet critical driving scenarios are severely underrepresented in real-world data. 
Existing solutions including safety-critical scenario generation and closed-loop learning often rely on rule-based heuristics, resampling methods and generative models learned from offline datasets, limiting their ability to produce diverse and novel challenges. While recent works leverage Vision Language Models (VLMs) to produce scene descriptions that guide a separate, downstream model in generating hazardous trajectories for agents, such two-stage framework constrains the generative potential of VLMs, as the diversity of the final trajectories is ultimately limited by the generalization ceiling of the downstream algorithm.
To overcome these limitations, we introduce \textbf{VILTA} (\textbf{V}LM-\textbf{I}n-the-\textbf{L}oop  \textbf{T}rajectory \textbf{A}dversary), a novel framework that integrates a VLM into the\textbf{\textit{ closed-loop training}} of AD agents. Unlike prior works, \methodname\ actively participates in the training loop by comprehending the dynamic driving environment and strategically generating challenging scenarios through direct, fine-grained editing of surrounding agents' future trajectories. This \textit{\textbf{direct-editing}} approach fully leverages the VLM's powerful generalization capabilities to create a diverse curriculum of plausible yet challenging scenarios that extend beyond the scope of traditional methods. We demonstrate that our approach substantially enhances the safety and robustness of the resulting AD policy, particularly in its ability to navigate critical long-tail events.
\end{abstract}


\begin{figure}[t!]
    \centering
    \includegraphics[width=1\linewidth]{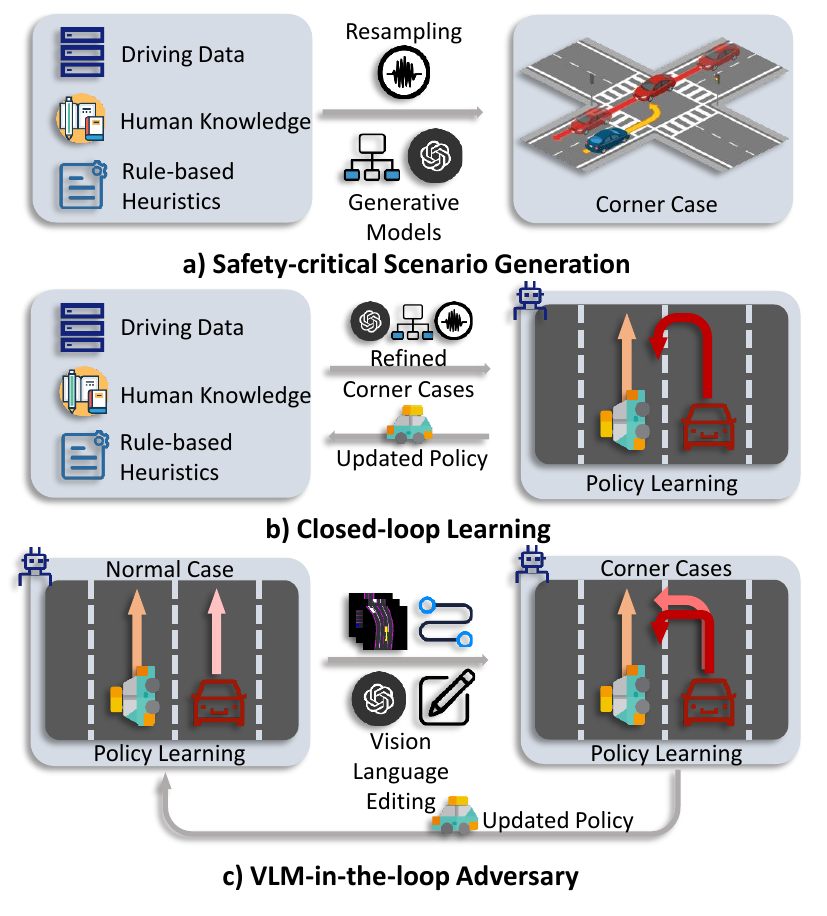}
    \caption{Comparison of different approaches to handling long-tail problem in autonomous driving. (a) Safety-critical scenario generation, which lacks use of the generated scenarios in training; (b) Closed-loop learning, which struggles to generate diverse scenarios; (c) Our proposed VLM-in-the-loop adversary, which is capable of generating scenarios that are both challenging and diverse.}
    \label{fig:intro}
\end{figure}

\section{Introduction}
The rapid advancement of artificial intelligence has catalyzed a paradigm shift in the automotive industry, accelerating the development of autonomous driving (AD) systems. In recent years, significant technological breakthroughs have been achieved across the entire AD stack, from perception and prediction to motion planning and end-to-end learning models~\cite{bevreview, predictionreview, planningreview, e2ereview}. These innovations have steadily enhanced the safety and efficiency of autonomous vehicles, bringing the prospect of fully autonomous transportation closer to reality.

Despite this remarkable progress, the long-tail distribution of real-world driving data~\cite{CoR, datasetsurvey}, a fundamental and persistent challenge of autonomous driving, impedes AD's widespread, reliable deployment. 
The vast majority of data collected from real-world driving consists of common, uneventful scenarios, such as highway cruising or following vehicles in moderate traffic. 
Conversely, safety-critical and complex situations called ``corner cases" are inherently rare. Such data imbalance leads to a critical performance gap: while AD models excel in handling nominal driving conditions~\cite{egomlp}, their ability to navigate rare and hazardous events remains underdeveloped and is a primary source of safety concerns.

To mitigate risks associated with the long-tail problem, researchers have created challenging offline datasets and online environments for testing and validation of pretrained driving models~\cite{diffscene, challenger, huang2024versatile, huang2024cadre, rowe2025scenario, lin2025causal, wang2021advsim}, as depicted in Fig.~\ref{fig:intro}a. Pioneering studies~\cite{chatscene, seektocollide, curricuvlm, zhang2025drivegen} utilized vast knowledge embedded within Vision-Language Models (VLMs) to generate high-level textual descriptions of a scene to adjust the priors of a separate, downstream generation framework. Such indirect, two-stage approaches fail to harness the full potential of VLMs, particularly their powerful and fine-grained generative capabilities.
Fig.~\ref{fig:intro}b shows a different line of works where dynamically generated challenging scenarios are integrated into the policy training loop~\cite{CAT, curricuvlm, CLIC}. However, these generated scenarios still rely on rule-based heuristics, resampling methods and generative models learned from existing datasets, which lacks the broad generality to create scenarios.

To address these limitations, we introduce a novel framework, \textbf{\textit{VILTA}}, which directly integrates a multimodal large model into the closed-loop training of autonomous driving agents for enhancing driving policy robustness, as shown in Fig.~\ref{fig:intro}c. In our proposed method, the VLM actively participates in the training loop by first comprehending the fine-grained representation of dynamic driving environment and then strategically generating challenging scenarios. This is achieved by having the VLM edit the future trajectories of surrounding agents to create scenarios that are both plausible and difficult for the ego-vehicle to handle. 

By doing so, our method makes two key contributions: 
\begin{itemize}
    \item Our framework fully leverages VLM's ability to produce a diverse and challenging curriculum of safety-critical scenarios integrated with policy learning.
    \item We introduce a highly efficient generation mechanism, where direct trajectory editing by the VLM provides a targeted way to craft adversarial interactions.  Experiments demonstrate that this approach improves the safety and robustness of the resulting autonomous driving policy, especially in its ability to navigate long tail events. Empirical analysis confirms that trajectories generated via VLM editing are more challenging in nature compared to those generated directly.
\end{itemize} 
\begin{figure*}[t]
    \centering
    \includegraphics[width=1\linewidth]{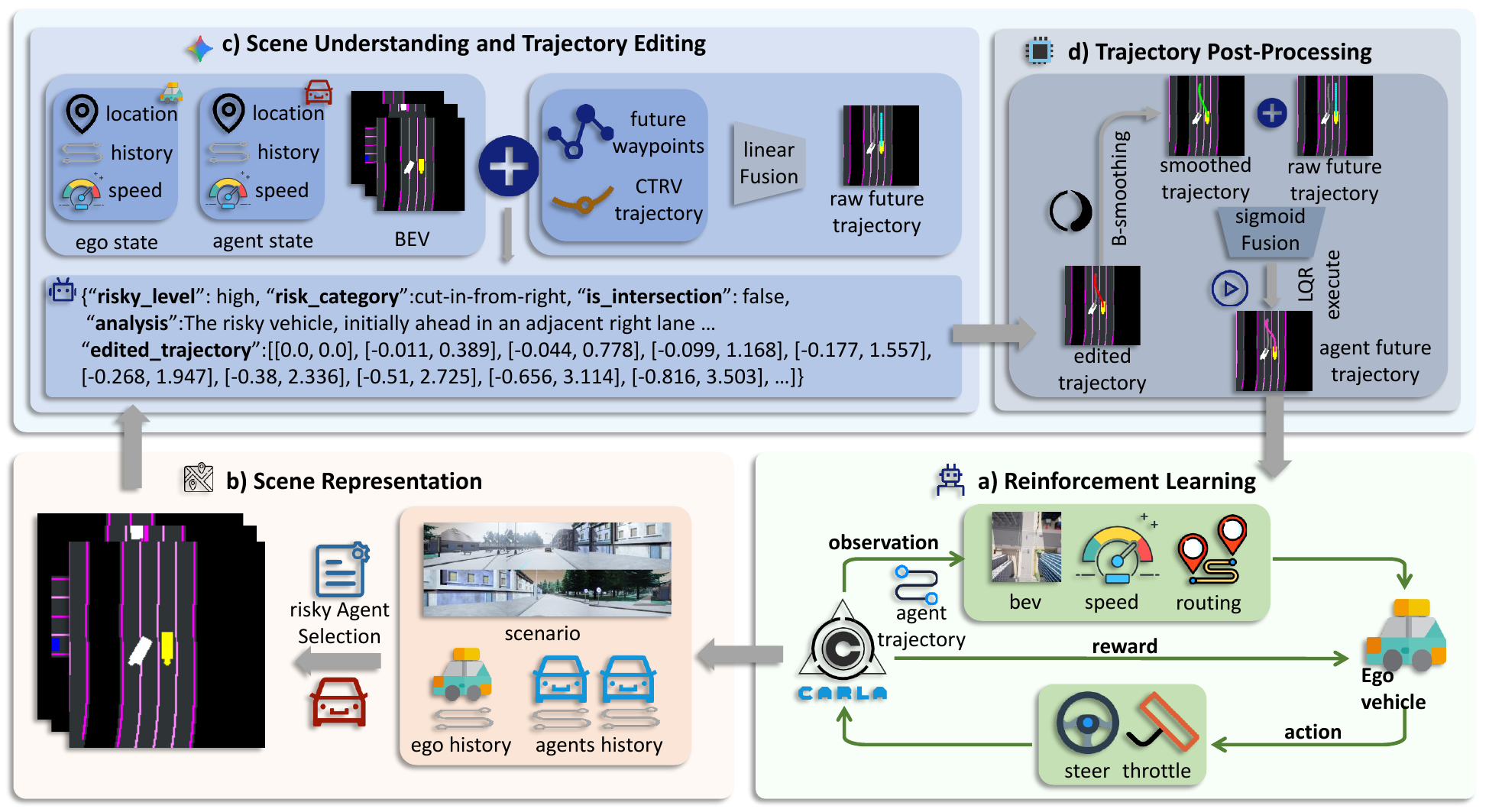}
    \caption{Overview of \methodname\ framework. (a) Reinforcement learning environment executes the edited trajectory of the risky agent, facilitates the training of the ego agent, and provides the initial scene representation; (b) The raw scene data is processed into a specific representation formatted for input to VLM; (c) Gemini performs both scene understanding and trajectory editing in a single pass, outputting a raw edited trajectory; (d) The post-processing stage ensures that the final trajectory is both smooth and kinematically feasible.}
    \label{fig:main}
\end{figure*}
\section{Related Work}

\subsection{Safety-Critical Scenario Generation}
Generating diverse and complex safety-critical scenarios in simulation is essential for evaluating autonomous driving (AD) systems, bypassing the high costs of real-world testing~\cite{drl}. Existing generation approaches include several main categories~\cite{scenariogenerationsurvey}. Data-driven methods~\cite{wheeler2015initial, wu2020spatiotemporal} suffer from the imbalanced nature of real-world data. Knowledge-based methods~\cite{diffscene} use predefined rules, which limits scenario diversity and generalizability. Adversarial generation~\cite{kuutti2020training} can produce diverse scenarios but fails to leverage real-world driving data.

More recently, Foundation Models (FMs) have been applied for their powerful generalization capabilities~\cite{bommasani2021opportunities, fmsurvey, gao2025words}. For instance, Chatscene~\cite{chatscene} translates language instructions into scenarios, CrashAgent~\cite{crashagent} converts crash reports into simulations. However, these FM-based methods typically adopt a two-stage approach: an FM first describes a scenario, and a downstream module then performs the generation. This indirect process fails to fully exploit the FM's generalization power and can be inefficient.

\subsection{Closed-loop Learning for Autonomous Driving}
Closed-loop learning is increasingly applied to AD to enhance model robustness by training on progressively challenging situations~\cite{adversariallearning}. Generating evolving curricula through environmental design can enhance the robustness of reinforcement learning (RL) agents~\cite{parker2022evolving}. Notable works include CAT~\cite{CAT}, which uses a closed-loop adversarial framework with real-world data, and SDM~\cite{SDM}, which models vehicle interactions as a Stackelberg game. CurricuVLM~\cite{curricuvlm} further integrates Vision-Language Models (VLMs) to analyze an AV's weaknesses and create personalized training curricula.
However, it still operates within a two-stage framework, using VLM's comprehension ability to guide the resampling of scenarios from an existing dataset. In contrast, our work breaks this two-stage paradigm, enabling VLM to utilize both its understanding and generative capabilities to directly craft complex scenarios, significantly improving the diversity and novelty of the generated challenges.


\section{Preliminaries}
\subsection{Partially Observable Markov Decision Process}

A Partially Observable Markov Decision Process (POMDP)~\cite{POMDP} extends the Markov Decision Process (MDP)~\cite{MDP} to formalize sequential decision-making under state uncertainty. A POMDP is defined by a tuple $(\mathcal{S}, \mathcal{A}, \mathcal{T}, \mathcal{R}, \Omega, \mathcal{O}, \gamma)$. After an agent takes an action $a \in \mathcal{A}$ in a state $s \in \mathcal{S}$, the environment transitions via $\mathcal{T}(s'|s,a)$ and yields a reward $\mathcal{R}(s,a,s')$. The agent receives an observation $o \in \Omega$ based on the probability $\mathcal{O}(o|s',a)$ instead of the true state. The objective is to find an optimal policy $\pi(a|s)$ that maximizes the expected discounted return $E\left[\sum_{t=0}^{\infty} \gamma^t \mathcal{R}(s_t, a_t, s_{t+1}) | \pi\right]$, where $\gamma$ is the discount factor.

\subsection{Problem Formulation}
Our objective is to train the autonomous vehicle by alternately exposing it to both complex and nominal scenarios, denoted as $\mathcal{S}_C, \mathcal{S}_N$, respectively. Based on this training approach, the objective is formulated in Eq.~\ref{obj} as follows:
\begin{equation}\label{obj}
  \pi^* = \arg\max_{\pi} E_{\substack{a_t\sim \pi(a|s)\\\mathcal{S}\in\{\mathcal{S}_N,\mathcal{S}_C,\dots\}\\s_{t+1}\sim \mathcal{T}(s'|s,a, \mathcal{S})}}\left[\sum_{t=0}^{\infty} \gamma^t \mathcal{R}(s, a, s') | \pi\right]
\end{equation},
where $\mathcal{T}(s_{t+1}|s_t,a_t, \mathcal{S})$ denotes state transition in scenario $\mathcal{S}$.
The notation $\mathcal{S}\in\{\mathcal{S}_N,\mathcal{S}_C,\mathcal{S}_N,\mathcal{S}_C,\dots\}$ indicates that the scenario $\mathcal{S}$ is drawn alternately from $\mathcal{S}_N$ and $\mathcal{S}_C$.

\section{Methodology}

The overall framework of our proposed method is illustrated in Fig.~\ref{fig:main}. We choose to use Gemini-2.5-Flash~\cite{comanici2025gemini} as our VLM, which is a publicly accessible API. 
This section begins by detailing the overall training framework for \methodname.
Subsequently, we introduce the representation for scenarios, and then describe each of the core components in sequence: scene understanding, trajectory editing, and trajectory post-processing.

\subsection{VLM In-the-loop Trajectory Adversary}

As depicted in Fig.~\ref{fig:main}, VILTA operates as a closed-loop training framework. During the training process, the RL environment provides state information about the current scene, which is then consolidated into a scene representation that is fed as input to the VLM. The VLM simultaneously performs scene understanding and trajectory editing. Finally, the resulting edited trajectory undergoes post-processing and is used to control the behavior of the ``risky agent" within the RL environment, thus closing the loop.

\subsection{Scene Representation}\label{scenerep}
To comprehensively capture scene information, we first structure the scene into a Bird's-Eye-View (BEV) representation as shown in Fig.~\ref{fig:main} (b).
Next, the vehicle closest to the ego vehicle within a predefined circle hazardous zone is selected as the agent responsible for generating the critical scenario. We then identify this agent's driving mode relative to the ego vehicle and assign a corresponding hazardous maneuver based on the selection rule as detailed in Tab.~\ref{tab:hm}.

\subsection{Scene Understanding}
Given the scene's BEV representation, the agent's driving mode relative to the ego vehicle, and the specified hazardous maneuver, the VLM outputs a structured understanding and analysis of the scene, as depicted in Fig.~\ref{fig:main}c.
VLM leverages its scene understanding capabilities to assess the current risk level and evaluate the appropriateness of the intended hazardous maneuver, using this analysis to produce the final edited trajectory.
In addition, VLM performs a check to determine if the ego vehicle is located at an intersection. If so, greater flexibility is permitted in the subsequent trajectory editing.

\subsection{Trajectory Editing}
\begin{table}[t]
  \centering
  \small
  
  \setlength{\tabcolsep}{2pt}
  \begin{tabular}{cccc|c}
  \toprule
    \makecell{Driving \\ Direction} & \makecell{Driving \\ Lane} & \makecell{Longitudinal \\ Position} & \makecell{Horizontal \\ Position} & \makecell{Hazardous \\ Maneuver}
    \\ 
  \midrule
    \multirow{4}{*}{Same} & \multirow{2}{*}{Same}  & Front & \multirow{2}{*}{--} & Sudden-brake\\
     & & Rear & & Overtake\\
     & \multirow{2}{*}{Different} & \multirow{2}{*}{--} & Left & Cut-in-left\\
     & & &Right & Cut-in-right\\
  \midrule
    Opposite & -- & -- & -- & \makecell{Lane-encroachment\\U-turn}\\
    
  \bottomrule
  \end{tabular}
  \caption{Agent's hazardous maneuver selection rule.}
  \label{tab:hm}
\end{table}
To ensure usability and generalizability, we leverage the pre-trained VLM without any fine-tuning to generate trajectories. A direct consequence, however, is that future trajectories generated directly via the standard Vision-Language-Action (VLA)~\cite{rt2} paradigm would lack the challenging nature, as shown in Sec.~\ref{empirical_anaylysis}.

To address this challenge, we instead employ a Vision-Language-Editing (VLE) paradigm. 
The inspiration for this approach is drawn from editing methodologies in the field of image manipulation, where a model can alter an image to adopt a specific style while preserving its underlying structure~\cite{meng2021sdedit, zhang2025scaling}.
We hypothesize that a similar phenomenon exists in the generation processes of large models, where editing raw information can preserve certain underlying structures. Applying this premise to the problem of challenging trajectory generation, our goal is to edit an initial, normal trajectory to render it more difficult, while simultaneously retaining its original overall motion trend.
In this approach, the VLM does not generate the trajectory from scratch; rather, it edits an initial trajectory for the agent which is produced by a rule-based method.
More specifically, we employ a weighted fusion method that combines the trajectory output from a Constant Turn Rate and Velocity (CTRV) model with predefined waypoints on the map. Details can be found in Alg.~\ref{alg:fuse}.
The generated trajectory is predominantly influenced by the CTRV model during its initial phase and by the map waypoints towards its conclusion. 
This baseline trajectory provides a stable foundation that effectively grounds the VLM's generation process, guiding it to produce feasible results and focusing its power on the editing task.
By editing this fused trajectory (denoted as $T_{base}$), we enhance the realism and reliability of the VLM's final output.

\begin{algorithm}[t]
\caption{Trajectory Fusion via Linear Weighting}\label{alg:fuse}
\begin{algorithmic}[1]
  \Require
  \Statex
    $T_{model}$: Trajectory from CTRV model.
    \Statex $T_{map}$:  Map's waypoints.
    \Statex $N$: Output trajectory length.

  \Ensure 
    $T_{base}$: The final fused trajectory.

  \State $T_{base} \gets [\ ]$ \texttt{// Initialize an empty list for the fused trajectory}
  
  \For{$i = 1$ to $N$}


    \State $p_{fused} \gets (1-\frac{i}{N}) \cdot T_{model}[i] + \frac{i}{N} \cdot T_{map}[i]$
    \State Append $p_{fused}$ to $T_{base}$
  \EndFor
  
  \State \Return $T_{base}$
\end{algorithmic}
\end{algorithm}

\subsection{Trajectory Post-processing}
While the design of the editing module enhances the reliability of the VLM's output trajectory, it does not in itself guarantee that the trajectory is kinematically feasible.
To address this, we introduce a post-processing module comprising three components. 
We denote the output trajectory from the editing module as $T_{edit}$.
First, B-spline smoothing~\cite{bsplinesmooth} is applied to the raw edited trajectory $T_{edit}$, yielding the smoothed trajectory $T_{B}$.
The second component is sigmoid fusion, where the smoothed trajectory $T_B$ is combined with the rule-based fused trajectory $T_{base}$ using a sigmoid weighting scheme:
\begin{equation}
\begin{aligned}
  w_i = \frac{1}{1 + \exp{(\frac{M(2i-N)}{N})}},\ i=1,2,\dots,N \\
  T_{curve}[i] = w_i \cdot T_{base}[i] + (1-w_i) \cdot T_{B}[i]
\end{aligned}
\end{equation}
where $N$ is the trajectory length and $M$ is the weighting factor.
The weight $w_i$ is initially high for $T_{base}$ to ensure behavioral continuity and prevent abrupt turns, and rapidly decreases to shift the trajectory's reliance towards the smoothed path.
Finally, the third component uses a Linear-Quadratic Regulator (LQR) controller~\cite{lqr}, taking the agent's kinematics to execute $T_{curve}$ and generate the final, kinematically plausible result $T_{final}$.

\section{Experiments}
\begin{table*}[t]
  
  \centering
  \small
  
  \begin{tabular}{ccccccccc}
  \toprule
    Town & Envrionment & Model & \makecell{Route\\Completion}$\uparrow$& \makecell{Total\\Distance}$\uparrow$ & \makecell{Crash\\Rate}$\downarrow$ & \makecell{Collision Per\\Kilometer}$\downarrow$ & \makecell{Collision\\Speed}$\downarrow$ & \makecell{Average\\Speed}$\uparrow$ \\
  \midrule
    \multirow{6}{*}{01} & \multirow{3}{*}{Challenging} & VLM-RL &0.80$\pm$0.03&4341.78$\pm$104.41&0.37$\pm$0.06&1.66$\pm$0.44&1.95$\pm$0.25&22.85$\pm$0.99 \\
    && CAT &0.70$\pm$0.06&3715.67$\pm$294.55&0.50$\pm$0.10&2.49$\pm$0.79&3.65$\pm$1.30&23.84$\pm$0.15\\
    && \methodname &0.93$\pm$0.09&5197.11$\pm$639.33&0.13$\pm$0.15&0.67$\pm$0.63&0.29$\pm$0.17&22.94$\pm$1.14 \\
    \cmidrule{2-9}
    &\multirow{3}{*}{Normal} & VLM-RL &0.88$\pm$0.10&4758.34$\pm$690.13&0.20$\pm$0.17&0.82$\pm$0.98&1.00$\pm$3.25&23.13$\pm$0.65\\
    && CAT &0.85$\pm$0.05&4912.2$\pm$134.19&0.33$\pm$0.15&1.45$\pm$0.78&1.99$\pm$0.25&23.03$\pm$0.28\\
    && \methodname &0.92$\pm$0.04&5156.23$\pm$208.23&0.13$\pm$0.06&0.96$\pm$0.57&2.17$\pm$1.57&22.73$\pm$0.56\\
  \midrule
    \multirow{6}{*}{02} & \multirow{3}{*}{Challenging} & VLM-RL &0.68$\pm$0.03&1441.66$\pm$121.46&0.57$\pm$0.06&18.86$\pm$3.39&4.87$\pm$0.95&18.96$\pm$2.11 \\
    && CAT &0.58$\pm$0.01&1076.81$\pm$17.70&0.73$\pm$0.06&22.44$\pm$0.44&6.61$\pm$0.79&21.87$\pm$0.12\\
    && \methodname &0.77$\pm$0.05&1588.91$\pm$157.71&0.50$\pm$0.10&18.13$\pm$7.36&3.63$\pm$3.13&20.13$\pm$2.74\\
    \cmidrule{2-9}
    &\multirow{3}{*}{Normal} & VLM-RL &0.86$\pm$0.06&1746.99$\pm$163.41&0.23$\pm$0.06&4.83$\pm$1.21&1.27$\pm$0.97&21.91$\pm$0.75 \\
    && CAT &0.89$\pm$0.06&1861.00$\pm$178.35&0.30$\pm$0.10&2.47$\pm$0.94&1.44$\pm$0.91&21.69$\pm$1.25\\
    && \methodname &0.91$\pm$0.08&1884.12$\pm$228.64&0.27$\pm$0.15&2.47$\pm$2.47&2.83$\pm$3.14&20.25$\pm$0.60\\
  \midrule
    \multirow{6}{*}{03} & \multirow{3}{*}{Challenging} & VLM-RL &0.62$\pm$0.08&2532.29$\pm$462.51&0.50$\pm$0.10&31.22$\pm$6.83&8.36$\pm$2.11&21.83$\pm$0.31 \\
    && CAT &0.62$\pm$0.02&2569.61$\pm$149.02&0.47$\pm$0.06&42.57$\pm$2.01&7.41$\pm$0.08&20.72$\pm$0.04\\
    && \methodname &0.68$\pm$0.06&2862.55$\pm$270.66&0.40$\pm$0.10&29.82$\pm$7.01&7.89$\pm$2.92&21.91$\pm$1.06\\
    \cmidrule{2-9}
    &\multirow{3}{*}{Normal} & VLM-RL &0.80$\pm$0.03&3461.05$\pm$183.96&0.33$\pm$0.06&12.36$\pm$0.66&4.08$\pm$2.49&23.61$\pm$0.27 \\
    && CAT &0.90$\pm$0.01&3545.09$\pm$44.65&0.23$\pm$0.06&0.75$\pm$0&0.33$\pm$0.14&22.26$\pm$0.84\\
    && \methodname &0.87$\pm$0.04&3787.55$\pm$171.41&0.23$\pm$0.06&11.00$\pm$0.75&2.17$\pm$0.25&22.15$\pm$0.55\\
  \midrule
  \multirow{6}{*}{Total} & \multirow{3}{*}{Challenging} & VLM-RL &2.10&8305.73&1.44&51.74&15.18&63.64 \\
    && CAT &1.90&7362.09&1.70&67.5&17.67&\textbf{66.43}\\
    && \cellcolor{gray!20} \methodname &\cellcolor{gray!20}\textbf{2.38}&\cellcolor{gray!20}\textbf{9648.57}&\cellcolor{gray!20}\textbf{1.03}&\cellcolor{gray!20}\textbf{48.62}&\cellcolor{gray!20}\textbf{11.81}&\cellcolor{gray!20}64.98\\
    \cmidrule{2-9}
    & \multirow{3}{*}{Normal} & VLM-RL &2.54&9966.38&0.76&18.01&6.35&\textbf{68.65} \\
    && CAT &2.64&10318.29&0.86&\textbf{4.67}&\textbf{3.76}&66.98\\
    && \cellcolor{gray!20} \methodname &\cellcolor{gray!20}\textbf{2.70}&\cellcolor{gray!20}\textbf{10827.90}&\cellcolor{gray!20}\textbf{0.63}&\cellcolor{gray!20}14.43&\cellcolor{gray!20}7.17&\cellcolor{gray!20}65.13\\
  \bottomrule
  \end{tabular}
  \caption{Performance comparison with several baselines. Averaged over 3 random seeds. Best results are marked in \textbf{bold}.}
  \label{tab:main}
\end{table*}
\subsection{Experimental Setups}
\subsubsection{Environments}

We conduct our experiments in the CARLA simulator~\cite{carla}. All models are trained on the Town02 map and evaluated on Town01, Town02, and Town03 using the final checkpoint during training. Following VLM-RL~\cite{vlmrl}, each map is populated with 20 autopilot vehicles, one of which is designated as a ``risky agent" per the methodology in Sec.~\ref{scenerep}. For each episode, the ego vehicle navigates between a randomly sampled start and destination, following the shortest path computed by the A* algorithm.

\subsubsection{RL Setting}

Our state representation follows VLM-RL~\cite{vlmrl}, concatenating an ego-centric BEV semantic map, the ego vehicle's state (steering, throttle, speed), and navigation information from 15 upcoming waypoints. The action space is a 2D continuous vector representing the steering angle in [-1, 1] and a combined throttle/brake value, where positive and negative values control throttle and brake, respectively.
For detailed information on the RL models and parameters used, please refer to the supplementary materials.

To steer the agent towards safe and efficient driving behaviors, we designed a comprehensive reward function, $R_{\text{total}}$, which is a weighted sum of three key components: a driving style reward ($R_{\text{style}}$), a vehicle-following reward ($R_{\text{follow}}$), and a safety penalty ($P_{\text{safety}}$). The total reward for a given state is calculated as:
\begin{equation}
  R_{\text{total}} = \alpha R_{\text{style}} + \beta R_{\text{follow}} - \gamma P_{\text{safety}}
  \label{eq:total_reward}
\end{equation}
where $\alpha$, $\beta$, and $\gamma$ are hyperparameters that balance the trade-offs between these objectives. 
\begin{itemize}
  \item \textbf{Driving Style Reward ($R_{\text{style}}$)} encourages smooth, stable, and centered driving. Following VLM-RL~\cite{vlmrl}, it is a product of four normalized factors, ensuring the agent must perform well across all aspects to achieve a high reward:
  \begin{equation}
    R_{\text{style}} = r_{\text{speed}} \cdot r_{\text{center}} \cdot r_{\text{angle}} \cdot r_{\text{stable}}
    \label{eq:style_reward}
  \end{equation}
  Here, $r_{\text{speed}}$ rewards maintaining a velocity within a predefined target range. $r_{\text{center}}$ is a function of the vehicle's lateral distance from the lane centerline. $r_{\text{angle}}$ penalizes angular deviation between the vehicle's heading and the lane's direction. Finally, $r_{\text{stable}}$ promotes stability by penalizing high variance in the vehicle's lateral position over a recent time window.

  \item \textbf{Following Reward ($R_{\text{follow}}$)} incentivizes car-following behavior by encouraging ego vehicle to maintain a safe, speed-dependent safe distance from the front vehicle:
\begin{equation}
     R_{\text{follow}}(d, v) = \begin{cases}
    \text{clip}(\frac{d - d_{\text{danger}}}{d_{\text{opt}}(v) - d_{\text{danger}}}, 0, 1) &\text{if } d \le d_{\text{opt}}(v) \\
    \frac{d_{\text{opt}}(v)}{d} &\text{if } d > d_{\text{opt}}(v)
\end{cases} 
\end{equation}
where $d$ is the distance to front vehicle, $d_{\text{danger}}$ is the dangerous distance threshold, and $d_{\text{opt}}(v)$ is the optimal following distance considering ego speed $v$.

  \item \textbf{Safety Penalty ($P_{\text{safety}}$)} serves as a general collision avoidance mechanism. It is a penalty term that increases sharply as the distance to the nearest surrounding vehicle (in any direction) falls below a critical safety threshold.
\end{itemize}

\subsubsection{Evaluation Metrics}
To provide a comprehensive assessment of the autonomous vehicle's performance, we evaluate both its safety and efficiency using a suite of metrics. Driving progression is measured by the average speed (\textbf{AS}, km/h), route completion rate (\textbf{RC}), and total distance (\textbf{TD}, m) traveled per episode. Safety is assessed based on the overall collision rate (\textbf{CR}), collision speed (\textbf{CS}, km/h), and collisions per kilometer (\textbf{CPM}).

\subsubsection{Baselines}
Our method is compared with two baselines:
\begin{itemize}
  \item \textbf{VLM-RL~\cite{vlmrl}:} A unified framework that combines VLMs with RL for safe autonomous driving by generating reward signals from image observations and contrasting language features from CLIP~\cite{CLIP}.
  \item \textbf{CAT~\cite{CAT}:} A framework that continuously improves the safety of autonomous driving agents by training them on safety-critical scenarios dynamically generated from pretrained motion prediction model. 
\end{itemize}

\subsection{Performance Comparison}
Our experiments are designed to answer the following two key questions:
\begin{itemize}
    \item Can \methodname\ agent maintain safe and efficient driving performance when confronted with challenging scenarios?
    \item Does \methodname\ suffer from ``catastrophic forgetting", where its performance on normal scenarios degrades after training on challenging ones?
\end{itemize}

We present a detailed evaluation of proposed \methodname\ against several baselines as shown in Tab.~\ref{tab:main}. 
For these main comparisons, \methodname\ utilizes a default 1:2 scenario alternation ratio. Note that this is a baseline configuration; as we will discuss in Sec.~\ref{sec:abla}, further tuning this frequency can yield even superior robustness.
The behavior of the surrounding vehicles differs by scenario type: in ``normal" scenarios, they operate under the control of a standard autopilot. In ``challenging" scenarios, they are dynamically controlled by the \methodname\ editing module to execute hazardous maneuvers targeted at the ego vehicle.
The ego vehicle is tasked with following a predefined navigation route. 
All models are trained in Town 02 and evaluated in Town 01-03 in both challenging and normal scenarios.

Beginning with the challenging testing scenarios, \methodname\ exhibits superior performance across all tested Towns. In terms of safety, it achieved the highest route completion rate, exceeding VLM-RL and CAT by 13.3\% and 25.3\% in total, respectively.
Simultaneously, \methodname\ records the lowest crash rate, representing a reduction of 28.5\% and 39.4\% compared to VLM-RL and CAT. 
Also, \methodname\ exhibits lowest average impact velocity, clearly indicating its enhanced safety profile. Furthermore, from an efficiency perspective, its average speed remained comparable to the leading baseline, CAT, showing only a marginal decrease of 2.2\%. This demonstrates that the significant safety gains were not achieved at the cost of efficiency.

Turning to the normal testing scenarios, the statistics reveal that \methodname\ performs similarly to the SOTA baselines. Specifically, \methodname\ achieved the best results for route completion rate, total distance traveled, and crash rate. In terms of efficiency, its performance was only marginally lower (by 5.1\%) than the top-performing baseline, VLM-RL.
These findings provide strong evidence that our training method, which alternates between challenging and normal scenarios, effectively mitigates catastrophic forgetting, allowing \methodname\ to preserve a strong balance of both safety and efficiency in standard driving conditions.

\subsection{Ablation Study}\label{sec:abla}
\begin{figure*}[h]
    \centering
    \includegraphics[width=1\linewidth]{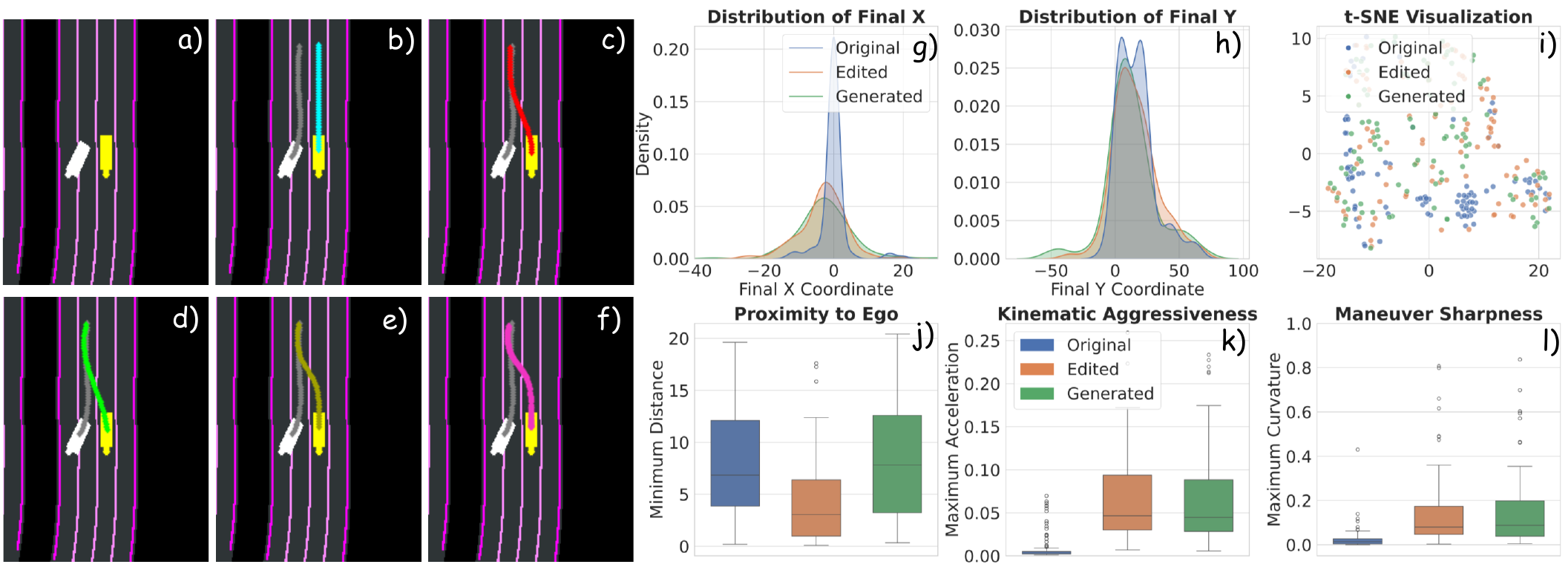}
    \caption{Trajectory visualization and empirical analysis. Panels (a-f) provide trajectory visualizations; panels (g-i) visualize the diversity of the trajectory distributions; and panels (j-l) present box plots of key trajectory features.}
    \vspace{-1mm}
    \label{fig:traj_ana}
\end{figure*}
To validate the effectiveness of each component in our proposed method, we conducted the following ablation studies:
\begin{itemize}
    \item W/o Post-Processing (w/o PP): The risky agent is controlled directly by the raw trajectory generated from VLM without any subsequent refinement.
    \item W/o Following Reward (w/o FR): The model is trained without the reward component $R_{\text{follow}}$ designed to encourage route adherence in Eq.~\ref{eq:total_reward}.
    \item W/o Vision-Language-Editing (w/o VLE): The risky agent is controlled by trajectories generated from VLM, without using VLE paradigm.
    \item Different Alternation Frequency between Normal and Challenging Scenarios: This ablation study investigates the effect of varying the frequency of challenging scenarios during RL training across three conditions: one challenging scenario is presented per every 2 (origin), 4, 8 or 16 normal scenarios, briefly denoted as x2, x4, x8 and x16 respectively.
\end{itemize}

\begin{table}[t]
    \centering
    \small
    
    \begin{tabular}{ccccc}
    \toprule
        Env. & Model & RC$\uparrow$ & CR$\downarrow$ & CPM$\downarrow$ \\
        \midrule
        \multirow{7}{*}{C} & w/o PP &0.73$\pm$0.03 &0.60$\pm$0 &18.75$\pm$2.09 \\
        &w/o FR&0.74$\pm$0.19&0.60$\pm$0.26&19.28$\pm$8.34 \\
        & w/o VLE &0.75$\pm$0.04&0.53$\pm$0.06&18.50$\pm$1.04 \\
        \cmidrule{2-5}
        &x2&0.77$\pm$0.05&0.50$\pm$0.10&18.13$\pm$7.36 \\
        &x4&0.85$\pm$0.03&\textbf{0.27$\pm$0.06}&11.21$\pm$14.89 \\
        &x8&\textbf{0.87$\pm$0.01}&0.30$\pm$0.10&\textbf{5.33$\pm$2.27} \\
        &x16&0.85$\pm$0.08&0.30$\pm$0.10&5.56$\pm$3.12\\
        \midrule
        \multirow{7}{*}{N} & w/o PP &0.89$\pm$0.05 &0.20$\pm$0.10 &2.24$\pm$0.50 \\
        &w/o FR&0.80$\pm$0.05&0.37$\pm$0.06&4.88$\pm$0.22 \\
        & w/o VLE &0.91$\pm$0.07&0.31$\pm$0.06&2.60$\pm$0.88\\
        \cmidrule{2-5}
        &x2&0.91$\pm$0.08&0.27$\pm$0.15&2.47$\pm$2.47 \\
        &x4&\textbf{0.94$\pm$0.05}&0.20$\pm$0.10&\textbf{1.30$\pm$0.85} \\
        &x8&0.92$\pm$0.05&0.23$\pm$0.15&2.30$\pm$1.62 \\
        &x16&0.91$\pm$0.07&\textbf{0.17$\pm$0.06}&1.60$\pm$0.88\\
    \bottomrule
    \end{tabular}
    \caption{Ablation study results. Averaged over 3 random seeds. ``C" and ``N" are brief denotations of ``Challenging" and ``Normal". ``x2" denotes the default setting used in our main experiments, while ``x4", ``x8", ``x16" explore sensitivity to frequency.}
    \label{tab:ablation}
\end{table}

As shown in Tab.~\ref{tab:ablation}, the inclusion of post-processing improves the route completion rate in challenging scenarios by 5.5\% while reducing CPM by 3.3\%. Adding the ``Following Reward" component boosts the route completion rate by 4.1\% and decreases CPM by 6.0\%.
The ablation of ``w/o VLE" likewise leads to a decline in performance on all evaluated metrics.
This is because the trajectories generated by our VLE paradigm and post-processing module are kinematically feasible, ensuring that alternating between ``hazardous" and ``normal" scenarios during training does not disrupt the physical consistency of the simulation environment.

Regarding the scenario alternation frequency, the results indicate that performance degrades when challenging scenarios are introduced either too infrequently (a 1:16 challenging-to-normal ratio) or too frequently (a 1:2 ratio). This suggests that finding an optimal balance is a critical empirical consideration for the training process. In our experiments, a frequency of one challenging scenario per eight normal ones (an 8:1 ratio) yielded the best trade-off. This suggests that while our main experiments utilized the x2 configuration, \methodname's performance could be further elevated by fine-tuning the scenario mixing ratio.

\subsection{Empirical Analysis}\label{empirical_anaylysis}

In this section, we present both qualitative and quantitative analyses of the trajectories before and after the editing process. 
Fig.~\ref{fig:traj_ana} a-f illustrate the detailed pipeline of our editing process, with the yellow and white vehicles representing the ``risky agent" and the ego vehicle, respectively. Initially, the scene representation includes the road structure and past trajectories (a). A raw future trajectory is then generated by linearly fusing CTRV model's prediction with map waypoints and get $T_{base}$ (b, cyan line). This raw trajectory is subsequently edited by the VLM (c, $T_{edit}$, red line) and then smoothed using a B-spline (d, $T_B$, green line). A sigmoid curve fusion stage then blends the smoothed trajectory from (d) with the raw future trajectory from (b) to produce an intermediate path (e, $T_{curve}$, yellow line). Finally, an LQR controller executes this path to produce the final, kinematically plausible trajectory (f, $T_{final}$, pink line).

Fig.~\ref{fig:traj_ana} g-l provide a quantitative analysis of 100 randomly sampled cases to verify the increased challenge and diversity of the edited trajectories (\textbf{Edited}) compared with the original agent future trajectories ($T_{base}$, \textbf{Original}) and trajectories directly generated from VLM without using our VLE paradigm ($T_{gen}$, \textbf{Generated}). 
Fig.~\ref{fig:traj_ana}g and ~\ref{fig:traj_ana}h show that the endpoint distribution of the final trajectories ($T_{final}$) is more dispersed than the initial ones ($T_{base}$). In Fig.~\ref{fig:traj_ana}i, we use t-SNE~\cite{tsne} to visualize a set of trajectory features (length, average speed, max curvature, endpoint), where the resulting 2D distribution for $T_{final}$ is significantly broader. The directly generated trajectories $T_{gen}$ exhibit a similar level of distributional diversity as the edited trajectories.

Fig.~\ref{fig:traj_ana} j-l plot the distributions for minimum distance to the ego vehicle, acceleration, and steering angle. 
In Fig.~\ref{fig:traj_ana}j, the minimum distance to the ego vehicle for the directly generated trajectories is comparable to that of the original ones, which indicates that the direct generation method merely enhances diversity without increasing the level of challenge to the ego vehicle.
Meanwhile, the edited trajectories $T_{final}$ feature smaller minimum distances as well as more extreme accelerations (Fig.~\ref{fig:traj_ana}k) and steering angles (Fig.~\ref{fig:traj_ana}l), all with wider distributions. Taken together, these results statistically confirm that our method enhances both the diversity and the challenging nature of the generated trajectories.

\section{Conclusions and Limitations}
This paper presented \methodname, a novel VLM-in-the-loop training framework for autonomous driving. By strategically integrating VLM into the training loop to perform fine-grained trajectory editing, \methodname\ generates a diverse curriculum of challenging scenarios. Experiments show that \methodname\ significantly enhances policy robustness in hazardous situations, increasing route completion and reducing collisions without degrading performance in normal conditions. The results validate that direct VLM integration is a highly effective strategy for creating more resilient autonomous driving agents.
Despite its success, our work has several limitations. First, validation is confined to simulation, necessitating future real-world testing to bridge the ``sim-to-real" gap. Second, the framework's performance is tied to the capabilities of the underlying VLM, warranting investigation into model-specific biases. Third, our current implementation focuses on single-agent adversarial scenarios, while real-world critical events can involve multiple actors. Finally, the initial identification of threats relies on predefined rules, suggesting an opportunity for more dynamic, learning-based threat discovery in future work.
Moreover, we plan to extend our framework to End-to-End (E2E) autonomous driving paradigms, investigating its potential to enhance the robustness of integrated perception-planning architectures.

\section*{Acknowledgements}
This work was supported by National Natural Science Foundation of China (Grant no.62473224 and 62461160260), in part by National Key Research and Development Program of China under grant 2023YFE0205800.
\bibliography{main}

\newpage
\twocolumn[
    \begin{@twocolumnfalse}

        \LARGE{\centering \textbf{Supplementary Material:} \\[0.5em]
\methodname: A VLM-in-the-Loop Adversary for Enhancing Driving Policy Robustness\\[1em]}
    \end{@twocolumnfalse}
]
\section{Implementation Details and Experiment Setup}
\subsection{\methodname\ Experiments}

Our implementation is built upon Stable-Baselines3~\cite{stable-baselines3}, a widely-used, open-source reinforcement learning framework. We use Soft Actor-Critic (SAC)~\cite{SAC} as our base learning method. During the reinforcement learning process, for the image-based inputs in observation space, we employ a simple Convolutional Neural Network (CNN) to extract features. Similarly, for the vector-based inputs, a simple Multi-Layer Perceptron (MLP) is used for feature extraction. Both the Actor and Critic networks are modeled as simple Multi-Layer Perceptrons (MLPs). The details of the training parameters and model architecture are listed in Tab.~\ref{tab:hyper}. We use Gemini-2.5-Flash API as the VLM agent.

\subsection{Baselines}
\subsubsection{VLM-RL}
For this baseline experiment, we faithfully replicated the official open-source implementation from the VLM-RL~\cite{vlmrl} paper with no modifications. It is worth noting that, to ensure a fair comparison, we utilized the final checkpoint obtained at the end of training for testing, rather than the best checkpoint during training as used in the original paper, to maintain consistency with \methodname\ and CAT which are also evaluated using the final checkpoint.

\subsubsection{CAT}
CAT~\cite{CAT} was originally developed and evaluated in the MetaDrive~\cite{li2021metadrive} simulation environment. To adapt this method for our experiments in CARLA, we began by collecting a new dataset. This dataset consists of 1000 scenarios generated within CARLA, and for each scenario, we sampled a 10-second trajectory, corresponding to 150 data frames. Each frame in a sample includes the Bird's-Eye-View (BEV) image, along with the current positions, headings, and velocities of the ego vehicle and all surrounding vehicles.

The collected data is then preprocessed into samples, where each sample contains three components: the current BEV image, the ego vehicle's past 1-second trajectory, and the past 1-second trajectory of a selected surrounding vehicle. Using this input, a DenseTNT~\cite{gu2021densetnt} model is trained to predict the future 2-second trajectory of the selected vehicle, along with its final position and a confidence score. Subsequently, a customized reinforcement learning (RL) agent is trained following the methodology presented in the original CAT paper. To ensure a fair and direct comparison with our proposed \methodname\ method, we keep all other experimental conditions identical. This includes employing the same Soft Actor-Critic (SAC) algorithm, and using identical state spaces, action spaces, reward functions, and hyperparameters.

\subsection{Computing Resources}
All experiments were conducted on a server equipped with a single NVIDIA H20 GPU, running the Debian operating system. The main experiment required approximately 1.5 days of computation time and consumed around 24 GB of Video RAM (VRAM).

\section{Additional Experiment Results}

\subsection{CARLA}

\begin{table*}[h!]
	\centering

		\begin{tabular}{lc}
			\toprule
			\textbf{Hyper-parameter} & \textbf{Value}\\
			\midrule
			\rowcolor{gray!20}\multicolumn{2}{c}{\textbf{SAC Algorithm}} \\
			Learning rate & Schedule: $1 \times 10^{-4} \to 5 \times 10^{-7}$ \\
			Discount factor ($\gamma$) & 0.98 \\
			Target update rate ($\tau$) & 0.02 \\
			Replay buffer size & 100,000 \\
			Batch size & 256 \\
			Training frequency & 64 steps \\
			Gradient steps & 64 \\
			Learning starts & 10,000 steps \\
            Total Timesteps & 1,000,000 steps\\
			Entropy coefficient & Auto-tuning \\
			Action smoothing & 0.75 \\
			
			\midrule
			\rowcolor{gray!20}\multicolumn{2}{c}{\textbf{Network Architecture}} \\
			Features dimension & 256 \\
			MLP architecture (Actor \& Critic) & [500, 300] \\
			CNN feature extractor (For 3-channel input, e.g., RGB-BEV) & \begin{tabular}{@{}l@{}}
			                             \\
			                            $C(16, k=5, s=2) \rightarrow \text{ReLU}$ \\
			                            $C(32, k=3, s=2) \rightarrow \text{ReLU}$ \\
			                            $C(64, k=3, s=2) \rightarrow \text{ReLU}$ \\
			                            $C(128, k=3, s=2) \rightarrow \text{ReLU}$ \\
			                            $C(256, k=3, s=1) \rightarrow \text{ReLU}$ \\
			                            $\text{Flatten}$
			                        \end{tabular} \\

			\midrule
			\rowcolor{gray!20}\multicolumn{2}{c}{\textbf{Observation Space}} \\
			Image resolution & $80 \times 120$ \\
			Image inputs & RGB-BEV, Segmentation-BEV \\
			Vector inputs & Steer, throttle, speed, waypoints \\
			
			\bottomrule
		\end{tabular}
	   \caption{Hyperparameters}
	\label{tab:hyper}
\end{table*}
In this section, we show additional test results conducted in CARLA town01-05, where all models are trained in town02. The CARLA town maps are identical with the ones in VLM-RL. Results are shown in Tab.~\ref{tab:append_main}.

\subsection{NuScenes}
We also conducted experiments on the nuScenes~\cite{caesar2020nuscenes} dataset to further validate our method. Specifically, since nuScenes is an offline autonomous driving dataset, we first modified it into a simulation environment. In this environment, all non-vehicle agents were removed, and the surrounding vehicles were controlled by their originally recorded trajectories, meaning they do not interact with the ego vehicle. We then employed our VILTA algorithm to designate one of these vehicles as the 'risky agent' and subsequently edited its future trajectory to create a potential threat to the ego vehicle.
We trained the ego-vehicle agent using the Proximal Policy Optimization (PPO)~\cite{ppo} algorithm, following a curriculum with a 4:1 alternation ratio of normal-to-challenging scenarios. All other settings were kept identical to those used in the CARLA experiments to ensure consistency.

\begin{table*}[h!]
\centering
\begin{tabular}{l c c c c c c}
\toprule
\multirow{3}{*}{Model} &\multicolumn{3}{c}{Normal}&\multicolumn{3}{c}{Challenging}\\
\cmidrule(lr){2-4}\cmidrule(lr){5-7}
& Crash\multirow{2}{*}{$\downarrow$} & Off-Road\multirow{2}{*}{$\downarrow$} & Success\multirow{2}{*}{$\uparrow$} & Crash\multirow{2}{*}{$\downarrow$} & Off-Road\multirow{2}{*}{$\downarrow$} & Success\multirow{2}{*}{$\uparrow$} \\
&Rate&Rate&Rate&Rate&Rate&Rate\\
\midrule
BC & \textbf{0.01} & 0.11 & 0.88 & 0.38 & 0.11 & 0.51 \\
PPO & 0.06 & 0.17 & 0.77 & 0.38 & 0.17 & 0.46 \\
BC + PPO & 0.06 & 0.05 & 0.89 & 0.36 & 0.02 & 0.62 \\
\rowcolor{gray!20}\methodname & 0.05 & \textbf{0.02} & \textbf{0.93} & \textbf{0.34} & \textbf{0.01} & \textbf{0.65} \\
\bottomrule
\end{tabular}
\caption{NuScenes experiment results.}
\label{tab:nuscenes}
\end{table*}

Tab.~\ref{tab:nuscenes} shows the experiment results in our nuScenes environment. Here we compare \methodname\ with three baselines:
\begin{itemize}
    \item \textbf{BC:} Behavior cloning~\cite{BC} imitation learning.
    \item \textbf{PPO:} Proximal policy optimization reinforcement learning.
    \item \textbf{BC+PPO:} The agent is first pre-trained using BC and subsequently fine-tuned using PPO.
\end{itemize}
During the evaluation phase, the agent is tested under both 'normal' (surrounding agents drive according to nuScenes logs) and 'challenging' (one risky agent drives according to VLM edited trajectory) conditions. In each test, the policy model generates a sequence of actions for the ego vehicle over a 3-second horizon. These actions are then executed in the simulation to roll out the ego vehicle's future trajectory. Finally, this generated ego trajectory is evaluated against the ground-truth future trajectories of surrounding vehicles to compute three key metrics: the crash rate, off-road rate, and success rate.

The results presented in Tab.~\ref{tab:nuscenes} clearly show that \methodname\ significantly outperforms the baselines, especially in challenging scenarios. It notably enhances the ego vehicle's safety under these conditions, achieving considerable reductions in both the crash rate and the off-road rate compared to all baselines. Moreover, \methodname\ demonstrates robustness against catastrophic forgetting, as its performance in normal scenarios remains on par with, and in some cases superior to, that of the baselines.

\begin{table*}[h!]
  
  \centering
  \small
  
  \begin{tabular}{ccccccccc}
  \toprule
    Town & Envrionment & Model & RC$\uparrow$ & TD$\uparrow$ & CR$\downarrow$ & CPM$\downarrow$ & CS$\downarrow$ & AS$\uparrow$ \\
  \midrule
    \multirow{6}{*}{01} & \multirow{3}{*}{Challenging} & VLM-RL &0.80$\pm$0.03&4341.78$\pm$104.41&0.37$\pm$0.06&1.66$\pm$0.44&1.95$\pm$0.25&22.85$\pm$0.99 \\
    && CAT &0.70$\pm$0.06&3715.67$\pm$294.55&0.50$\pm$0.10&2.49$\pm$0.79&3.65$\pm$1.30&23.84$\pm$0.15\\
    && \methodname &0.93$\pm$0.09&5197.11$\pm$639.33&0.13$\pm$0.15&0.67$\pm$0.63&0.29$\pm$0.17&22.94$\pm$1.14 \\
    \cmidrule{2-9}
    &\multirow{3}{*}{Normal} & VLM-RL &0.88$\pm$0.10&4758.34$\pm$690.13&0.20$\pm$0.17&0.82$\pm$0.98&1.00$\pm$3.25&23.13$\pm$0.65\\
    && CAT &0.85$\pm$0.05&4912.2$\pm$134.19&0.33$\pm$0.15&1.45$\pm$0.78&1.99$\pm$0.25&23.03$\pm$0.28\\
    && \methodname &0.92$\pm$0.04&5156.23$\pm$208.23&0.13$\pm$0.06&0.96$\pm$0.57&2.17$\pm$1.57&22.73$\pm$0.56\\
  \midrule
    \multirow{6}{*}{02} & \multirow{3}{*}{Challenging} & VLM-RL &0.68$\pm$0.03&1441.66$\pm$121.46&0.57$\pm$0.06&18.86$\pm$3.39&4.87$\pm$0.95&18.96$\pm$2.11 \\
    && CAT &0.58$\pm$0.01&1076.81$\pm$17.70&0.73$\pm$0.06&22.44$\pm$0.44&6.61$\pm$0.79&21.87$\pm$0.12\\
    && \methodname &0.77$\pm$0.05&1588.91$\pm$157.71&0.50$\pm$0.10&18.13$\pm$7.36&3.63$\pm$3.13&20.13$\pm$2.74\\
    \cmidrule{2-9}
    &\multirow{3}{*}{Normal} & VLM-RL &0.86$\pm$0.06&1746.99$\pm$163.41&0.23$\pm$0.06&4.83$\pm$1.21&1.27$\pm$0.97&21.91$\pm$0.75 \\
    && CAT &0.89$\pm$0.06&1861.00$\pm$178.35&0.30$\pm$0.10&2.47$\pm$0.94&1.44$\pm$0.91&21.69$\pm$1.25\\
    && \methodname &0.91$\pm$0.08&1884.12$\pm$228.64&0.27$\pm$0.15&2.47$\pm$2.47&2.83$\pm$3.14&20.25$\pm$0.60\\
  \midrule
    \multirow{6}{*}{03} & \multirow{3}{*}{Challenging} & VLM-RL &0.62$\pm$0.08&2532.29$\pm$462.51&0.50$\pm$0.10&31.22$\pm$6.83&8.36$\pm$2.11&21.83$\pm$0.31 \\
    && CAT &0.62$\pm$0.02&2569.61$\pm$149.02&0.47$\pm$0.06&42.57$\pm$2.01&7.41$\pm$0.08&20.72$\pm$0.04\\
    && \methodname &0.68$\pm$0.06&2862.55$\pm$270.66&0.40$\pm$0.10&29.82$\pm$7.01&7.89$\pm$2.92&21.91$\pm$1.06\\
    \cmidrule{2-9}
    &\multirow{3}{*}{Normal} & VLM-RL &0.80$\pm$0.03&3461.05$\pm$183.96&0.33$\pm$0.06&12.36$\pm$0.66&4.08$\pm$2.49&23.61$\pm$0.27 \\
    && CAT &0.90$\pm$0.01&3545.09$\pm$44.65&0.23$\pm$0.06&0.75$\pm$0&0.33$\pm$0.14&22.26$\pm$0.84\\
    && \methodname &0.87$\pm$0.04&3787.55$\pm$171.41&0.23$\pm$0.06&11.00$\pm$0.75&2.17$\pm$0.25&22.15$\pm$0.55\\
  \midrule
    \multirow{6}{*}{04} & \multirow{3}{*}{Challenging} & VLM-RL &0.67$\pm$0.06&12166.54$\pm$307.06&0.33$\pm$0.06&15.45$\pm$5.89&2.97$\pm$1.44&20.52$\pm$0.59 \\
    && CAT &0.71$\pm$0.09&11477.43$\pm$1365.3&0.3$\pm$0.1&11.48$\pm$4.03&8.32$\pm$4.12&22.63$\pm$0.43\\
    && \methodname &0.77$\pm$0.05&13147.5$\pm$942.08&0.3$\pm$0.1&8.04$\pm$3.57&8.55$\pm$1.13&24.1$\pm$0.33\\
    \cmidrule{2-9}
    &\multirow{3}{*}{Normal} & VLM-RL &0.85$\pm$0.04&16116.64$\pm$218.6&0.17$\pm$0.06&0.43$\pm$0.38&0$\pm$0&21.99$\pm$0.44 \\
    && CAT &0.9$\pm$0.01&14457.06$\pm$207.21&0.13$\pm$0.06&65.22$\pm$1.36&1.43$\pm$0.06&21.88$\pm$0.04\\
    && \methodname &1$\pm$0&16802.18$\pm$264&0$\pm$0&0$\pm$0&0$\pm$0&24.37$\pm$0.82\\
  \midrule
    \multirow{6}{*}{05} & \multirow{3}{*}{Challenging} & VLM-RL &0.74$\pm$0.08&2618.39$\pm$547.74&0.47$\pm$0.12&3.43$\pm$0.93&2.95$\pm$0.43&24.55$\pm$0.63 \\
    && CAT &0.86$\pm$0.05&3235.73$\pm$184.63&0.27$\pm$0.06&3.12$\pm$2.32&2.41$\pm$1.08&23.39$\pm$0.36\\
 && \methodname &0.76$\pm$0.05&2554.14$\pm$209.03&0.43$\pm$0.06&3.51$\pm$1.13&3.3$\pm$1.69&23.08$\pm$0.68\\
\cmidrule{2-9}
&\multirow{3}{*}{Normal} & VLM-RL &0.77$\pm$0.08&2721.99$\pm$570.27&0.43$\pm$0.06&3.18$\pm$0.91&2.53$\pm$2.55&24.91$\pm$0.23 \\
&& CAT &0.99$\pm$0.02&3707.33$\pm$78.45&0.03$\pm$0.06&0.16$\pm$0.27&0.04$\pm$0.07&23.81$\pm$0.05\\
&& \methodname &0.76$\pm$0.04&2749.44$\pm$172.15&0.4$\pm$0.1&3.38$\pm$0.92&3.22$\pm$1.72&22.98$\pm$1.61\\
  \midrule
  \multirow{6}{*}{Total} & \multirow{3}{*}{Challenging} & VLM-RL &3.51&23090.66&2.24&70.62&\textbf{21.10}&108.71 \\
    && CAT &3.47&22075.25&2.27&82.1&28.40&\textbf{112.45}\\
    && \cellcolor{gray!20}{\methodname} &\cellcolor{gray!20}\textbf{3.91}&\cellcolor{gray!20}\textbf{25350.21}&\cellcolor{gray!20}\textbf{1.76}&\cellcolor{gray!20}\textbf{60.17}&\cellcolor{gray!20}23.66&\cellcolor{gray!20}112.16\\
    \cmidrule{2-9}
    &\multirow{3}{*}{Normal} & VLM-RL &4.16&28805.01&1.36&21.62&8.88&\textbf{115.55} \\
    && CAT &\textbf{4.53}&28482.7&\textbf{1.02}&70.05&\textbf{5.23}&112.67\\
    && \cellcolor{gray!20}\methodname &\cellcolor{gray!20}4.46&\cellcolor{gray!20}\textbf{30379.52}&\cellcolor{gray!20}1.03&\cellcolor{gray!20}\textbf{17.81}&\cellcolor{gray!20}10.39&\cellcolor{gray!20}112.48\\
  \bottomrule
  \end{tabular}
  \caption{Performance comparison in Town01-05. Averaged over 3 random seeds. Best results are marked in \textbf{bold}.}
  \label{tab:append_main}
\end{table*}

\section{Details of VLM Input/Output}\label{visualization_vlm}
In this section, we present a detailed visualization of the inputs and outputs for VLM.

\begin{tcolorbox}[breakable, colback=gray!10, colframe=gray!80, width=\linewidth ,title=A simplified version of the VLM input.]

As a professional autonomous driving scenario designer, your core task is to edit a future trajectory for a designated ‘risky’ vehicle to create a challenging interaction scenario with the ego vehicle. Your expertise involves editing the risky vehicle's trajectory by designing a challenging yet realistic interaction based on the provided scene, vehicle data, and road structure. You will integrate bird’s-eye-view (BEV) imagery—which visualizes the ego vehicle (white), risky agent (yellow), neutral vehicles (blue), drivable area (gray), and lane markings (magenta)—with vehicle dynamics data like position, trajectories, and speed for both the ego and risky vehicles. Based on inputs including BEV scale, simulation FPS, and vehicle states, you are to edit a specified number of future waypoints for the risky agent over a set time period. The required risky behavior depends on the initial `risk\_category`: an `opposite` position requires a u-turn or lane-encroachment; `same\_behind` an aggressive-overtake; `same\_ahead` a sudden-brake; and `adjacent\_left` or `adjacent\_right` a cut-in. If the ego is at an intersection, as denoted by a boolean flag, you have greater freedom to generate diverse risky trajectories based on the intersection structure; otherwise, you must follow the predefined rules. The generated trajectory must be kinematically feasible, smooth, stay within the drivable area, avoid collisions with neutral targets, and create a close-call or overlap with the ego’s future path. Your final output must be a single JSON object containing: a `risk\_level` (`High`, `Medium`, or `Low`); a `risk\_category` describing the maneuver; the `is\_intersection` boolean; a detailed `analysis` paragraph explaining your reasoning; and a list of the specified number of `waypoints` as `[relative\_x, relative\_y]` coordinates formatted to three decimal places.

\end{tcolorbox}

\begin{figure*}[h!]
    \centering
    \begin{tcolorbox}[title=VLM Output, myboxstyle]
    {\small
    \begin{verbatim}
    
{
    "response": {
        "risk_level": "High",
        "risk_category": "u-turn",
        "is_intersection": true,
        "analysis": "The ego vehicle is approaching an intersection from the right,
        with its future trajectory indicating it will proceed through the
        intersection, likely turning left. The risky agent is positioned at the bottom
        stem of the T-intersection, initially moving forward slightly. To create a
        challenging scenario, the risky agent will perform an aggressive U-turn
        (turning right from its perspective) immediately upon entering the
        intersection. This maneuver is designed to place the risky vehicle directly
        into the path of the ego vehicle as the ego attempts to navigate the
        intersection. Specifically, around frame 16 (approximately 1.06 seconds into
        the future), the risky vehicle will be at [5.000, -1.000], while the ego
        vehicle will be at [5.017, -0.305]. This creates a high-risk, direct conflict
        scenario where the risky vehicle cuts off the ego, forcing the ego to execute
        an emergency braking or evasive maneuver to prevent a collision. The
        trajectory is kinematically feasible, with speeds ranging between 7.5 m/s and
        9.375 m/s, and remains within the drivable area, avoiding other neutral
        vehicles.",
        "waypoints": [[0.0, 0.5], [0.0, 1.0], [0.0, 1.5], [0.0, 2.0], [0.5, 2.125],
        [1.0, 2.25], [1.5, 2.375], [2.0, 2.5], [2.5, 2.125], [3.0, 1.75], [3.5,
        1.375], [4.0, 1.0], [4.25, 0.5], [4.5, 0.0], [4.75, -0.5], [5.0, -1.0],
        [5.125, -1.5], [5.25, -2.0], [5.375, -2.5], [5.5, -3.0], [5.625, -3.5], [5.75,
        -4.0], [5.875, -4.5], [6.0, -5.0], [6.125, -5.5], [6.25, -6.0], [6.375, -6.5],
        [6.5, -7.0], [6.625, -7.5], [6.75, -8.0], [6.875, -8.5], [7.0, -9.0], [7.125,
        -9.5], [7.25, -10.0], [7.375, -10.5], [7.5, -11.0], [7.625, -11.5], [7.75,
        -12.0], [7.875, -12.5], [8.0, -13.0]]
    }
}

    
    \end{verbatim}
    }
    \end{tcolorbox}
    \caption{A complete sample of the output generated by the VLM.}
    \label{fig:vlm-output}
\end{figure*}

\section{Edited Results}\label{editres}
\begin{figure*}[h!]
    \centering
    \includegraphics[width=1\linewidth]{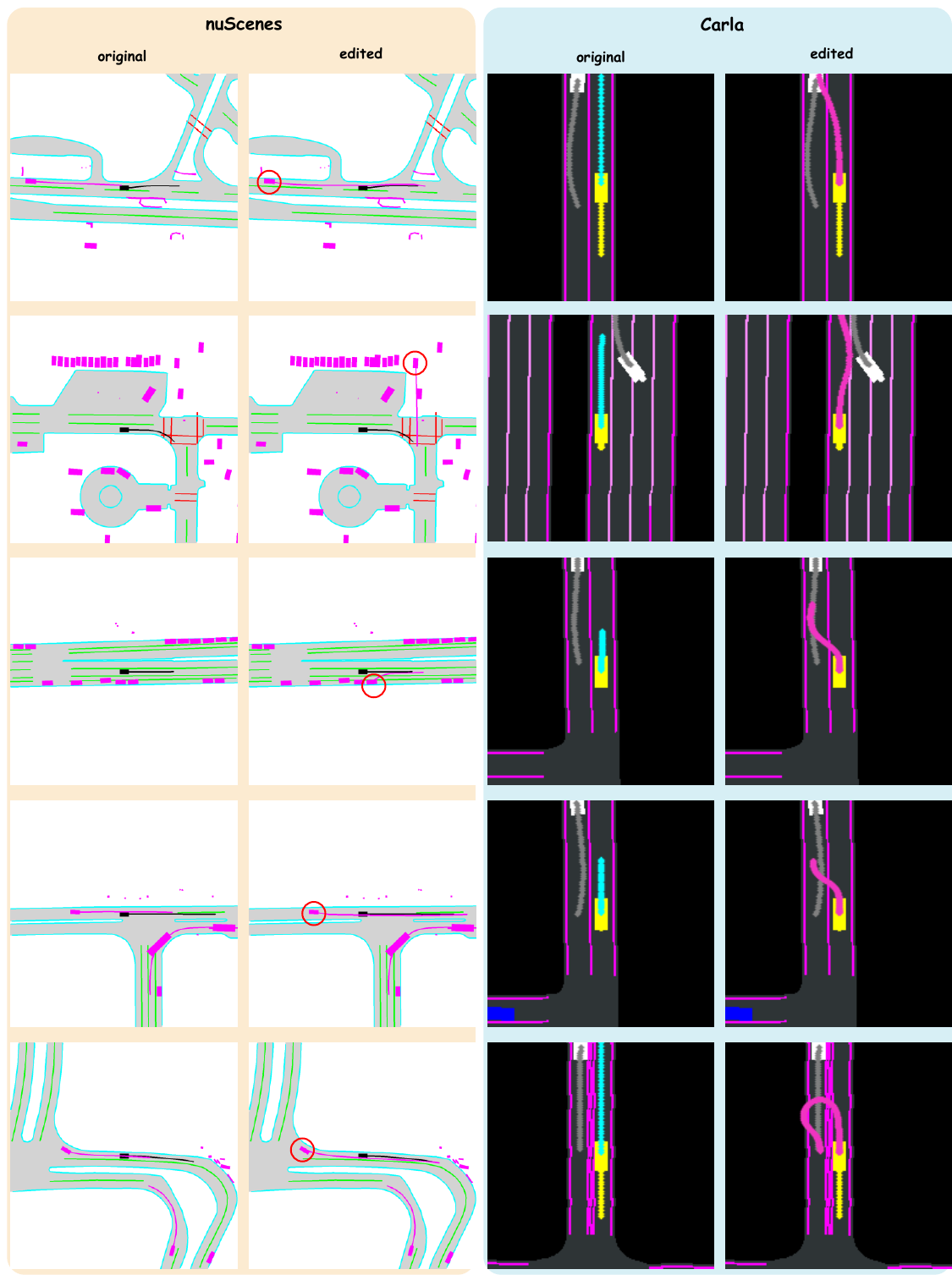}
    \caption{Edited trajectories visualization in both nuScenes and CARLA environments.}
    \label{fig:edit_res}
\end{figure*}
In this section, we present additional visualizations comparing the trajectories before and after the editing process.
Fig.~\ref{fig:edit_res} is organized into four columns to compare the trajectories before and after editing in both the nuScenes and CARLA environments:
\begin{itemize}
    \item Column 1: Depicts the original ground-truth trajectories from the nuScenes dataset, where the ego vehicle is shown in black and surrounding vehicles is shown in pink. 
    \item Column 2: Illustrates the same nuScenes scenario after our editing process, showing the clear alteration of the risky agent's path. The risky agent is circled in red.
    \item Column 3: Presents the original trajectories from the CARLA simulator. The ego vehicle is white, the surrounding vehicle is yellow, and their respective future paths are visualized in grey and cyan.
    \item Column 4: Shows the corresponding CARLA trajectory after editing, where the future path of the surrounding vehicle has been modified and is now colored pink.
\end{itemize}

\end{document}